\documentclass{article}

\usepackage[numbers]{natbib}

 \usepackage[final]{nips_2017}
\usepackage{subfigure}
\usepackage[utf8]{inputenc} 
\usepackage[T1]{fontenc}    
\usepackage{hyperref}       
\usepackage{url}            
\usepackage{booktabs}       
\usepackage{amsfonts}       
\usepackage{nicefrac}       
\usepackage{microtype}      
\usepackage{amsmath}
\usepackage{algorithm,algorithmic}
\usepackage{graphicx}
\title{Variational Deep Q Network}

\author{
 Yunhao Tang\\
  Department of IEOR\\
 Columbia University\\
  \texttt{yt2541@columbia.edu} \\
   \And
   Alp Kucukelbir \\
   Department of Computer Science \\
  Columbia University \\
  \texttt{alp@cs.columbia.edu} \\
}

\begin{document}

\maketitle

\begin{abstract}
We propose a framework that directly tackles the probability distribution of the value function parameters in Deep Q Network (DQN), with powerful variational inference subroutines to approximate the posterior of the parameters. We will establish the equivalence between our proposed surrogate objective and variational inference loss. Our new algorithm achieves efficient exploration and performs well on large scale chain Markov Decision Process (MDP).
\end{abstract}

\section*{Introduction}
Deep reinforcement learning (RL) has enjoyed numerous recent successes in video games, board games, and robotics control \cite{schulman2015,duanxi2016,levine2016,silver2016}. Deep RL algorithms typically apply naive exploration schemes such as $\epsilon-$greedy \cite{mnih2013,sutton1998}, directly injecting noise into actions \cite{timothy2016}, and action level entropy regularization \cite{williams1992}. However, such local perturbations of actions are not likely to lead to systematic exploration in hard environments \cite{fortunato2017}. Recent work on deep exploration \cite{osband2016} applies the bootstrap to approximate the posterior probability of value functions, or injects noise into value function/policy parameter space \cite{fortunato2017,lipton2016}. 

We propose a framework that directly approximates the distribution of the value function parameters in a Deep Q Network. We present a surrogate objective that combines the Bellman error and an entropy term that encourages efficient exploration. The equivalence between our proposed objective and variational inference loss allows for the optimization of parameters using powerful variational inference subroutines. Our algorithm can be interpreted as performing approximate Thompson sampling \cite{william1933}, which can partially justify the algorithm's efficiency. We demonstrate that the algorithm achieves efficient exploration with good performance on large scale chain Markov decision processes that surpasses DQN with $\epsilon-$greedy exploration \cite{mnih2013} and NoisyNet \cite{fortunato2017}.

\section{Background}

\subsection*{Markov Decision Process}

A Markov Decision Process is a tuple $(\mathcal{S},\mathcal{A},\mathcal{P},\mathcal{R},\rho)$ where we have state space $\mathcal{S}$, action space $\mathcal{A}$, transition kernel $\mathcal{P}: \mathcal{S} \times \mathcal{A} \mapsto \mathcal{S}$, reward function $\mathcal{R}: \mathcal{S}\times \mathcal{A} \mapsto \mathbb{R}$ and initial distribution $\rho$ over states. A policy is a mapping from state to action $\pi: \mathcal{S} \mapsto \mathcal{A}$. At time $t$ in state $s_t \in \mathcal{S}$, the agent takes action $a_t$, transitions to $s_{t+1}$ under $\mathcal{P}$, and receives reward $r_t$ under $\mathcal{R}$. Unless otherwise stated, the expectation over state $s_{t+1}$ and reward $r_t$ is with respect to transition kernel $\mathcal{P}$ and reward function $\mathcal{R}$. The objective is to find a policy $\pi$ to maximize the discounted cumulative reward 
$$\mathbb{E}_{s_0 \sim \rho, a_t \sim \pi(\cdot|s_t)} \big[ \sum_{t=0}^\infty r_t \gamma^t \big]$$
where $\gamma$ is a discount factor. Being in state $s$, the action-value function $Q^\pi(s,a)$ under policy $\pi$ is defined as the expected cumulative reward that could be received by first taking action $a$ and then following policy $\pi$
$$Q^\pi (s,a) = \mathbb{E}_{a_t \sim \pi(\cdot|s_t)} \big[ \sum_{t=0}^\infty r_t \gamma^t | s_0 = s, a_0 = a \big]$$

Under policy $\pi$ the the Bellman error under policy $\pi$ is
\begin{align}
J(\theta) = \mathbb{E}_{s_0 \sim \rho, a_t\sim \pi(\cdot|s_t)}\left[\left(Q_\theta(s_t,a_t) - \max_a \mathbb{E}\left[ r_t + \gamma Q_\theta(s_{t+1},a)\right] \right)^2\right]
\label{eq:bellman}
\end{align}

Bellman's optimality condition specifies that, for optimal policy $\pi^\ast$, its action-value function $Q^\ast(s,a)$ satisfies the following condition
$$Q^\ast(s_t,a_t) = \max_a \mathbb{E} \big[ r_t + \gamma Q^\ast(s_{t+1},a)\big]$$ for any $s_t\in \mathcal{S}$ and $a_t \in \mathcal{A}$. Hence the optimal action value function $Q^\ast(s,a)$ has zero Bellman error, any action value function with zero Bellman error is also optimal.

\subsection*{Deep Q Network}
Deep Q Networks (DQN) \cite{mnih2013} proposes to approximate the action value function $Q^\pi(s,a)$ by a neural network $Q_\theta(s,a)$ with parameters $\theta$. Let $\pi_\theta$ be greedy policy with respect to $Q_\theta(s,a)$. The aim is to choose $\theta$ such that the Bellman error of Equation (\ref{eq:bellman}) is minimized.

In practice, the expectation in (\ref{eq:bellman}) is estimated by $K$ sample trajectories collected from the environment, each assumed to have period $T$. Let $Q_\theta^{(i)}(s_t^{(i)},a_t^{(i)})$ be the approximate action value function computed at state-action pair $(s_t^{(i)},a_t^{(i)})$ on the $i$th sample trajectory. The approximate Bellman error is 
$$J(\theta) \approx \tilde{J}(\theta) = \frac{1}{K}\frac{1}{T}\sum_{i=1}^K \sum_{t=0}^{T-1} (\hat{Q}_\theta^{(i)}(s_t^{(i)},a_t^{(i)})-r_t-\gamma \max_a \hat{Q}_\theta^{(i)}(s_{t+1}^{(i)},a))^2$$
Equivalently, let $N = K \times T$ be total number of samples and $\{s_j,a_j,r_j,s_j^\prime\}$ be a relabeling of $\{s_t^{(i)},a_t^{(i)},r_t^{(i)},s_{t+1}^{(i)}\}$ by sample number. Then, the error can be written as 
\begin{align}
 \tilde{J}(\theta) = \frac{1}{N} \sum_{j=1}^N (Q_\theta(s_j,a_j) - r_j - \max_{a^\prime} Q_\theta(s_j^\prime,a^\prime))^2 
 \label{eq:bellmanerror}
 \end{align}
In (\ref{eq:bellmanerror}) the term $r_j + \max_{a^\prime}Q_\theta(s_j^\prime,a^\prime)$ is called target value. To minimize $\hat{J}(\theta)$ is essentially to minimize the discrepancy between target value and prediction $Q_\theta(s_j,a_j)$. To stabilize training, \cite{mnih2013} proposes to compute the target value by a target network with parameter $\theta^{-}$. The target network has the same architecture as the original network but its parameters are slowly updated, allowing the target distribution to be more stationary. The final approximate Bellman error is 
\begin{align}
\hat{J}(\theta) = \frac{1}{N} \sum_{j=1}^N (Q_\theta(s_j,a_j) - r_j - \max_{a^\prime} Q_{\theta^{-}}(s_j^\prime,a^\prime))^2
\label{eq:targetbellman}
\end{align}
The parameter $\theta$ is updated by stochastic gradient descent on the final approximate Bellman error $\theta \leftarrow \theta - \alpha \nabla_\theta \hat{J}(\theta)$ where $\alpha$ is the learning rate.

\subsection{Variational Inference}
Given a generative model with parameter $\theta$, the samples $X$ are generated from distribution $X \sim p(X|\theta)$. Define prior $p(\theta)$ on the parameters $\theta$. Given generated data $D = \{X_j\}_{j=1}^N$ the posterior of $\theta$ is computed by Bayes rule $p(\theta|D) = \frac{p(\theta,D)}{p(D)}$.

In most cases it is challenging to evaluate $p(\theta|D)$ directly. Consider using a variational family distribution $q_\phi(\theta)$ with parameter $\phi$ to approximate the posterior. One approach is to minimize the KL divergence between $q_\phi(\theta)$ and $p(\theta|D)$
\begin{align}
\min_\phi\ \mathbb{KL}(q_\phi(\theta)||p(\theta|D))
\label{eq:kldivergence}
\end{align}
In complex generative models such as Bayesian neural networks, we can approximately solve the above minimization problem using gradient descent. The gradient of (\ref{eq:kldivergence}) can be derived as an expectation, which is estimated by sample averages in practical implementation. When $\mathbb{KL}(q_\phi(\theta)||p(\theta|D))$ is approximately minimized, we could directly infer from $q_\phi(\theta)$ \cite{blei2017,blei2015,kucukelbir2016automatic}.

\section{Related Methods}
DQN \cite{mnih2013} is one of the first successful frameworks in deep reinforcement learning. Built upon the original work, there have been numerous attempts to improve the learning stability and sample efficiency, such as prioritized replay \cite{schaul2016}, double DQN \cite{hasselt2016} and duel DQN \cite{wang2016} among others.

The duality between control and inference \cite{todorov2008} encourages the application of variational inference to reinforcement learning problems. \cite{furmston2010} propose specialized inference techniques applicable to small MDPs yet they could not be scaled to large problems.

VIME (Variational Information Maximization Exploration) \cite{duan2016} proposes to encourage exploration by informational bonus. The algorithm learns a dynamics model of the environment and then computes informational bonus based on changes in the posterior distribution of the dynamics model. The informational bonus is computed from a variational family distribution that approximates the posterior. This offers an novel approach to exploration yet the exploration strategy is still intuitively local.

Bootstrapped DQN \cite{osband2015,osband2016} proposes to approximate the formidable posterior of value function parameters with bootstrap. Different heads of the Bootstrapped DQN are trained with different sets of bootstrapped experience data. Multiple heads of the Bootstrapped DQN entail diverse strategies and encourage exploration. Though bootstrapping can be performed efficiently by parallel computing, this method is in general computationally costly.

Recent work on NoisyNet \cite{fortunato2017} proposes to add noise to value function parameters or policy parameters directly. The true parameters of the model are parameters that govern the distribution of value function/policy parameters. By a re-parametrization trick, the distribution parameters are updated by conventional backpropagation. NoisyNet applies randomization in parameter space, which corresponds to randomization in policy space and entails more consistent exploration.

BBQ Network \cite{lipton2016} is closest in spirit to our work. BBQ Network also randomizes in policy space and  achieves good performance on dialogue tasks when combined with VIME \cite{duan2016} and an entire pipeline of natural language processing system. Compared to their work, our formulation starts from a surrogate objective that explicitly encourages exploration and establishes its connection with variational inference. The variational interpretation allows us to leverage efficient black box variational subroutines to update network parameters.

\section{Proposed Algorithm}

\subsection{Formulation}

As in the DQN formulation, the optimal action value function is approximated by a neural network $Q_\theta(s,a)$ with parameter $\theta$. Consider $\theta$ following a parameterized distribution $\theta\sim q_\phi(\theta)$ with parameter $\phi$. The aim is to minimize an expected Bellman error
$$\ \mathbb{E}_{\theta\sim q_\phi(\theta)} \big[ \sum_{j=1}^N (Q_\theta(s_j,a_j) - r_j - \max_{a^\prime} Q_\theta(s_{j}^\prime,a^\prime))^2\big]$$
where we have adopted the sample estimate of Bellman error (without $\frac{1}{N}$) as in (\ref{eq:bellmanerror}). The distribution $q_\phi(\theta)$ specifies a distribution over $\theta$ and equivalently specifies a distribution over policy $\pi_\theta$. To entail efficient exploration, we need $q_\phi(\theta)$ to be dispersed. Let $\mathbb{H}(\cdot)$ be the entropy of a distribution. Since large $\mathbb{H}(q_\phi(\theta))$ implies dispersed $q_\phi(\theta)$, we encourage exploration by adding an entropy bonus $-\mathbb{H}(q_\phi(\theta))$ to the above objective 
\begin{equation}
 \mathbb{E}_{\theta\sim q_\phi(\theta)} \big[ \sum_{j=1}^N (Q_\theta(s_j,a_j) - r_j - \max_{a^\prime} Q_\theta(s_{j}^\prime,a^\prime))^2\big] - \lambda \mathbb{H}(q_\phi(\theta))
 \end{equation}
where $\lambda > 0$ is a regularization constant, used to balance the expected Bellman error and entropy bonus. The aim is to find $\phi$ that achieves low expected Bellman error while encompassing as many different policies as possible. 

As in DQN \cite{mnih2013}, to stabilize training, we have a target parameter distribution $q_{\phi^{-}}(\theta^{-})$ over $\theta^{-}$ with slowly updated parameters $\phi^{-}$. The target $r_j + \max_{a^\prime} Q_\theta(s_{j}^\prime,a^\prime)$ is computed by a target network $\theta^{-}$ sampled from the target distribution $\theta^{-} \sim q_{\phi^{-}} (\theta^{-})$. The final surrogate objective is 
\begin{align}
 \mathbb{E}_{\theta\sim q_\phi(\theta),\theta^{-}\sim q_{\phi^{-}}(\theta^{-})} \big[ \sum_{j=1}^N (Q_\theta(s_j,a_j) - r_j - \max_{a^\prime} Q_{\theta^{-}}(s_{j}^\prime,a^\prime))^2\big] - \lambda \mathbb{H}(q_\phi(\theta))
 \label{eq:surrogate}
 \end{align}
 
\subsection{Variational Inference Interpretation}
Next, we offer an interpretation of minimizing surrogate objective (\ref{eq:surrogate}) as minimizing a variational inference loss. Let target value $d_j = r_j + \max_{a^\prime} Q_{\theta^-}(s_{j}^\prime,a^\prime)$ be given (computed by target network $\theta^-$) and let $\sigma = \sqrt{\frac{\lambda}{2}}$. The objective (\ref{eq:surrogate}) is equivalent up to constant multiplication to
\begin{align}
\mathbb{E}_{\theta\sim q_\phi(\theta)} \big[ \frac{1}{2\sigma^2}\sum_{j=1}^N (Q_\theta(s_j,a_j) - d_j))^2\big] - \mathbb{H}(q_\phi(\theta))
\label{eq:variationalloss}
\end{align}
To bridge the gap to variational inference, consider $Q_\theta(s,a)$ as a Bayesian neural network with parameter $\theta$ with improper uniform prior $p(\theta)\propto 1$. The network generates data with Gaussian distribution $d_j \sim \mathbb{N}(Q_\theta(s,a),\sigma^2)$ with given standard error $\sigma$. Given data $D = \{d_j\}_{j=1}^N$, $p(\theta|D)$ denotes the posterior distribution of parameter $\theta$. The above objective (\ref{eq:variationalloss}) reduces to KL divergence between $q_\phi(\theta)$ and the posterior of $\theta$ 
\begin{equation}
\mathbb{KL}\big[q_\phi(\theta)||p(\theta|D)\big]
\label{eq:kldivergenceloss}
\end{equation}
Hence to update parameter $\phi$ based on the proposed objective (\ref{eq:surrogate}) is equivalent to find a variational family distribution $q_\phi(\theta)$ as approximation to the posterior $p(\theta|D)$. In fact, from (\ref{eq:variationalloss}) we know that the posterior distribution $p(\theta|D)$ is the minimizer distribution of (\ref{eq:surrogate}), 

Here we have established the equivalence between surrogate objective (\ref{eq:surrogate}) and variational inference loss (\ref{eq:variationalloss}). In general, we only need to assume Gaussian generative model $Q_\theta(s,a)$ and any variational inference algorithm will perform approximate minimization of Bellman error. This interpretation allows us to apply powerful black-box variational inference packages, such as Edward \cite{tran2017}, to update the value function and leverage different black-box algorithms \cite{blei2015,kucukelbir2016automatic}. We can recover the original DQN as a special case of Variational DQN. See Appendix.

\subsection{Algorithm}
The variational inference interpretation of proposed objective allows us to leverage powerful variational inference machinery to update policy distribution parameter $\phi$. 

We have a principal distribution parameter $\phi$ and a target distribution parameter $\phi^{-}$. At each time step $t$, we sample $\theta \sim q_\phi(\theta)$ and select action by being greedy with respect to $Q_\theta(s_t,a)$. The experience tuple $\{s_t,a_t,r_t,s_{t+1}\}$ is added to a buffer $R$ for update. When updating parameters, we sample a mini-batch of tuples $\{s_j,a_j,r_j,s_j^\prime\}_{j=1}^N$ and compute target values $d_j = r_j + \max_{a^\prime} Q_{\theta^{-}}(s_j^\prime,a^\prime)$ using target network parameter $\theta^{-} \sim q_{\phi^{-}}(\theta^{-})$. Then we evaluate the KL divergence in (\ref{eq:kldivergenceloss}) using $d_j$ as generated data and improper uniform prior $p(\theta)$. The parameter $\phi$ is updated by taking one gradient descent step in KL divergence. The target parameter $\phi^{-}$ is updated once in a while as in the original DQN. The pseudocode is summarized below. The algorithm can be interpreted as performing approximate Thompson sampling \cite{william1933}. See Appendix.

\begin{algorithm}[H]
	\begin{algorithmic}[1]
		\STATE INPUT:  improper uniform prior $p(\theta)$; target parameter update period $\tau$	; learning rate $\alpha$; generative model variance $\sigma^2$
		\STATE INITIALIZE: parameters $\phi,\phi^{-}$; replay buffer $R \leftarrow \{\}$; step counter $counter \leftarrow 0$
		\FOR {$e=1,2,3...E$}
		\WHILE {episode not terminated}
		\STATE $counter \leftarrow counter + 1$
		\STATE Sample $\theta \sim q_\theta(\phi)$
		\STATE In state $s_t$, choose $a_t = \arg\max_a Q_\theta(s_t,a)$, get transition $s_{t+1}$ and reward $r_t$
		\STATE Save experience tuple $\{s_t,a_t,r_t,s_{t+1}\}$ to buffer $R$
		\STATE Sample $N$ parameters $\theta_j^{-} \sim q_\theta(\phi^{-})$ and sample $N$ tuples $D = \{s_j,a_j,t_j,s_j^\prime\}$ from $R$
		\STATE Compute target $d_j =  r_j + \max_{a^\prime} Q_{\theta_j^{-}}(s_j^\prime,a^\prime)$ for $j$th tuple in $D$ 
		\STATE Take gradient $\Delta \phi$ of the KL divergence in (\ref{eq:kldivergenceloss}) 
		\STATE $\phi\leftarrow \phi - \alpha \Delta \phi$
		\IF {$counter \ \text{mod}\  \tau = 0$}
		\STATE Update target parameter $\phi^{-} \leftarrow \phi$
		\ENDIF
		\ENDWHILE
		\ENDFOR
	\end{algorithmic}
	\caption{Variational DQN}
\end{algorithm}

\section{Testing Environments}
\subsection{Classic Control Tasks}
These four classic control tasks are from OpenAI Gym environments \cite{brockman2016}. They all require the agent to learn a good policy to properly control mechanical systems. Among them, \emph{MountainCar} and \emph{Acrobot} are considered as more challenging since to solve the environment requires efficient exploration. For example in  \emph{MountainCar}, a bad exploration strategy will get the car stuck in the valley and the agent will never learn the optimal policy.

\subsection{Chain MDP}
The chain MDP \cite{osband2016} (Figure 1) serves as a benchmark environment to test if an algorithm entails deep exploration. The environment consists of $N$ states and each episode lasts $N+9$ time steps. The agent has two actions $\{\text{left},\text{right}\}$ at each state $s_i,1\leq i\leq N$, while state $s_1,s_N$ are both absorbing. The transition is deterministic. At state $s_1$ the agent receives reward $r = \frac{1}{1000}$, at state $s_N$ the agent receives reward $r = 1$ and no reward anywhere else. The initial state is always $s_2$, making it hard for the agent to escape local optimality at $s_1$. 

If the agent explores randomly (assign $\frac{1}{2}$ probability to choose $\text{left}$ and $\text{right}$ respectively), the expected number of time steps required to reach $s_N$ is $2^{N-2}$. For large $N$, it is almost not possible for the randomly exploring agent to reach $s_N$ in a single episode, and the optimal strategy to reach $s_N$ by keeping choosing $\text{right}$  will never be learned.

\begin{figure}[h]
\centering
\subfigure[Chain MDP with $N$ states]{\includegraphics[width=.8\linewidth]{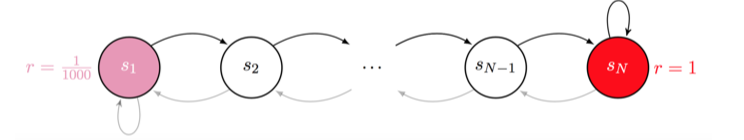}}
\caption{Illustration of Chain MDP \cite{osband2016}}
\end{figure}

The feature $\phi(s)$ of state $s$ is used as input to the neural network $Q_\theta(s,a)$ to compute approximate action value function. As suggested in \cite{osband2016}, we consider feature mapping $\phi_{\text{therm}}(s) = \mathbb{I}\{x\leq s\}$ in $\{0,1\}^N$. 
where $\mathbb{I}\{\cdot\}$ is the indicator function. 

\newpage
\section{Experiments}

\subsection{Classic Control Tasks}
We compare Variational DQN and DQN on these control tasks. Both Variational DQN and DQN can solve the four control tasks within a given number of iterations, yet they display quite different characteristics in training curves. On simple tasks like \emph{CartPole} (Figure 2 (a) and (b)), Variational DQN makes progress faster than DQN but converges at a slower rate. This is potentially because Variational DQN optimizes over the sum of Bellman error and exploration bonus, and the exploration bonus term in effect hinders fast convergence on optimal strategy for simple tasks.

\begin{figure}[h]
\centering
\subfigure[CartPole-v0]{\includegraphics[width=.4\linewidth]{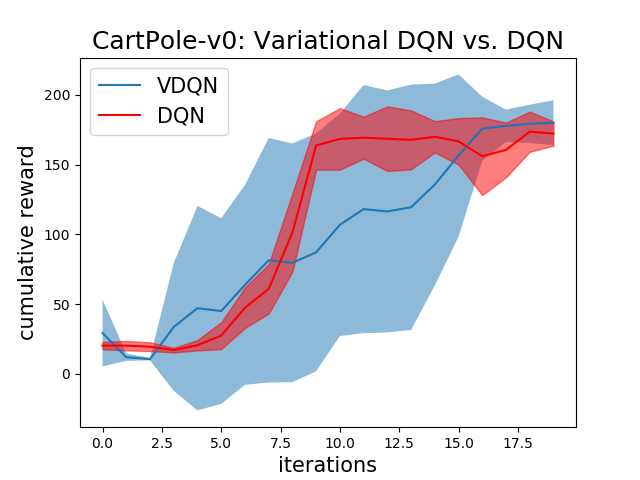}}
\subfigure[CartPole-v1]{\includegraphics[width=.4\linewidth]{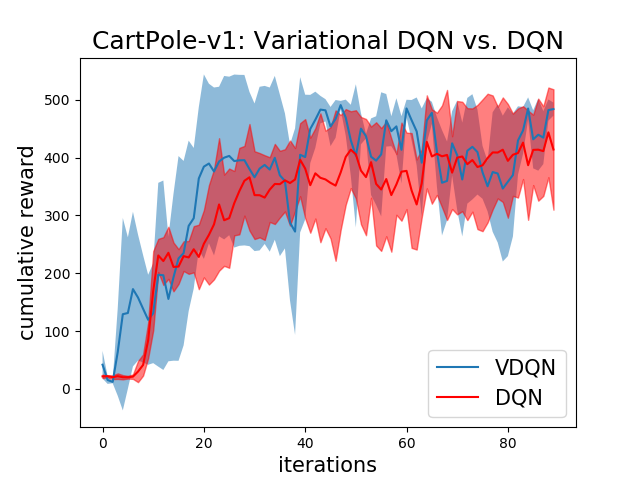}}
\subfigure[Acrobot-v1]{\includegraphics[width=.4\linewidth]{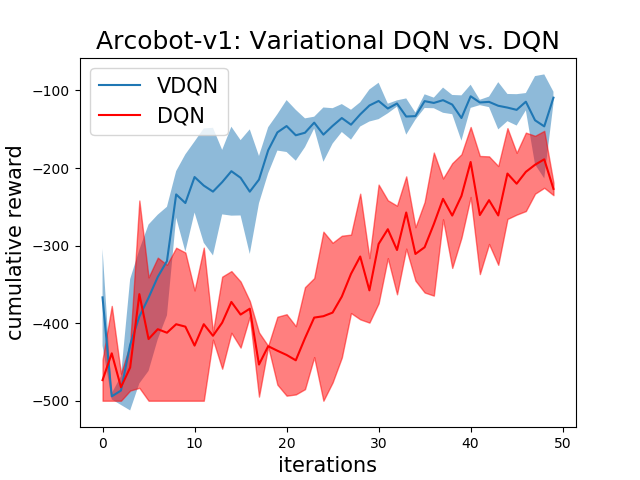}}
\subfigure[MountainCar-v0]{\includegraphics[width=.4\linewidth]{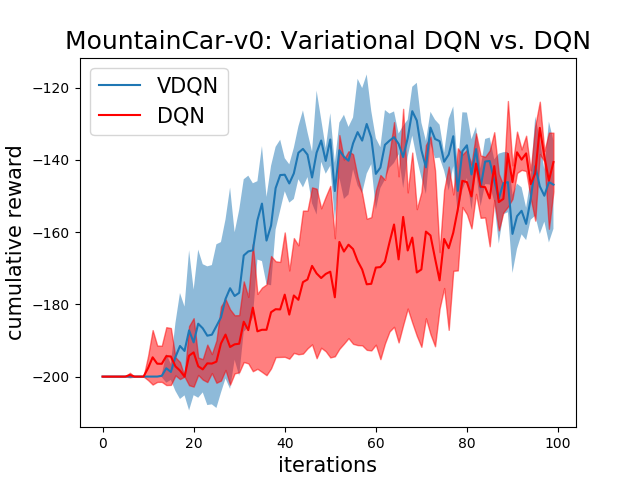}}
\caption{Variational DQP vs. DQN: Training curves of both algorithms on four control tasks. Training curves are averaged over multiple initializations of both algorithms. Each iteration is $10$ episodes.}
\end{figure}

\subsection{Chain MDP}
We compare Variational DQN, DQN and NoisyNet on Chain MDP tasks. For small $N\leq 10$, all algorithms converge to optimal policy within reasonable number of iterations, and even the training curves of Variational DQN and Noisy network are very similar. When $N$ increases such that $N\geq 50$, DQN barely makes progress and cannot converge to optimal policy, while NoisyNet converges more slowly to optimal policy and oscillates much. When $N\geq100$, both DQN and NoisyNet barely make progress during tranining.

The performance of Variational DQN is fairly stable across a large range of $N$. For $N\leq 70$, Variational DQN converges to optimal policy within 500 episodes (50 iterations) on average. However, when $N$ keeps increasing such that $N\geq 100$, Variational DQN takes longer time to find the optimal policy but it makes steady improvement over time.

The big discrepancy between the performance of these three algorithms on Chain MDP tasks is potentially due to different exploration schemes. As stated previously, under random exploration, the expected number of steps it takes to reach $s_N$ is approximately $2^{N-1}$. Since DQN applies $\epsilon-$greedy for exploration, for large $N$ it will never even reach $s_N$ within limited number of episodes, letting alone learning the optimal policy. NoisyNet maintains a distribution over value functions, which allows the agent to consistently execute a sequence of actions under different policies, leading to more efficient exploration. However, since NoisyNet does not explicitly encourage dispersed distribution over policies, the algorithm can still converge prematurely if the variance parameter converges quickly to zero. On the other hand, Variational DQN encourages high entropy over policy distribution and can prevent such premature convergence.

\begin{figure}[h]
\centering
\subfigure[Chain MDP $N=5$]{\includegraphics[width=.43\linewidth]{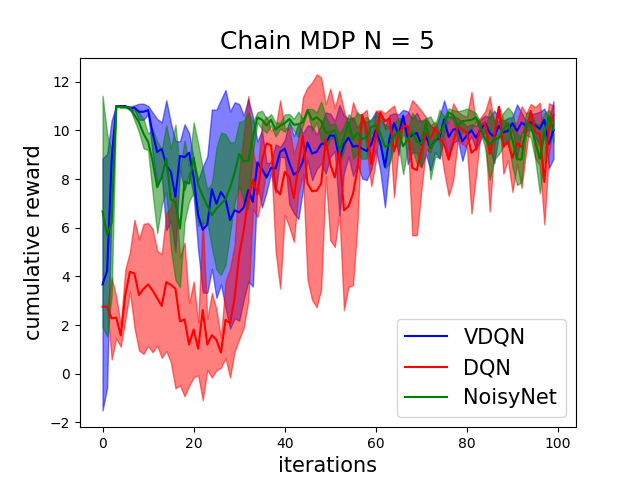}}
\subfigure[Chain MDP $N=10$]{\includegraphics[width=.43\linewidth]{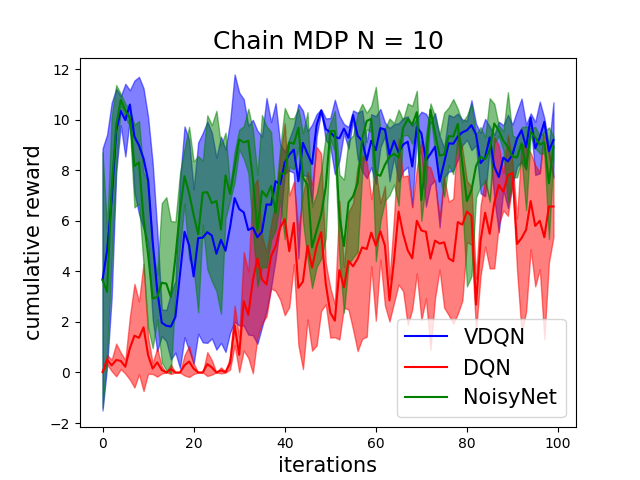}}
\subfigure[Chain MDP $N=50$]{\includegraphics[width=.43\linewidth]{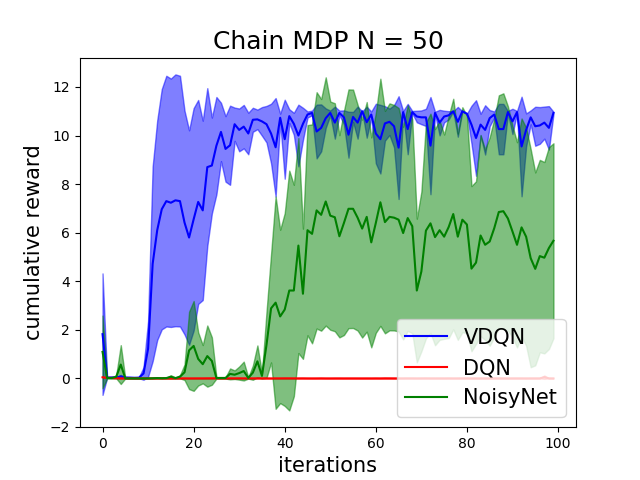}}
\subfigure[Chain MDP $N=100$]{\includegraphics[width=.43\linewidth]{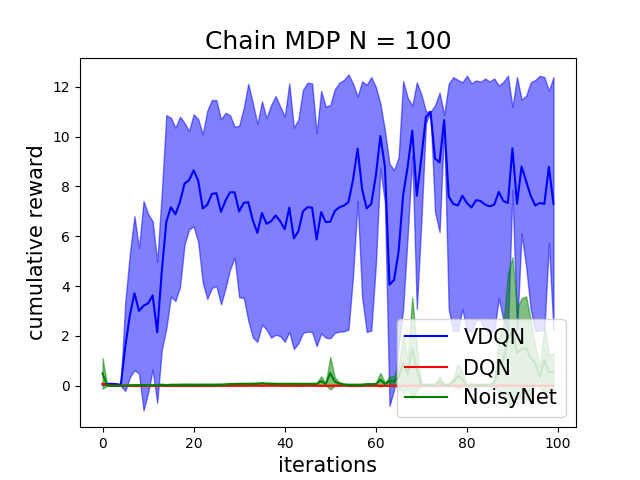}}
\caption{Variational DQP vs. DQN vs. NoisyNet in Chain MDP: Training curves of three algorithms on Chain MDP tasks with different $N$. Training curves are averaged over multiple initializations of both algorithms. Each iteration is $10$ episodes.}
\end{figure}

To further investigate why Variational DQN can do systematic and efficient exploration of the environment, we plot the state visit counts of Variational DQN and DQN for $N = 32$ in Figure 4. Let $c_n$ be the visit count to state $s_n$ for $1\leq n\leq N$. In each episode, we set $c_n = 1$ if the agent ever visits $s_n$ and $c_n = 0$ otherwise. The running average of $c_n$ over consecutive episodes is the approximate visit probability $p_n$ of state $s_n$ under current policy. In Figure 4 we show visit probability $p_n$ for $n=1$ (locally optimal absorbing state), $n=N$ (optimal absorbing state) and $n=\frac{N}{2}$. The probability $p_n$ for $n=\frac{N}{2}$ is meant to show if the agent ever explores the other half of the chain in one episode.

At early stage of training ($\text{iterations} \leq 10$), Variational DQN starts with and maintains a relatively high probability of visiting all three states. This enables the agent to visit $s_{N}$ for sufficient number of trials and converges to the optimal policy of keeping going to the right and reaching $s_N$. On the other hand, DQN occasionally has nontrivial probability of visiting $s_{\frac{N}{2}}$ due to $\epsilon-$greedy random exploration. But since DQN does not have enough momentum to consistently go beyond $s_{\frac{N}{2}}$ and visit $s_N$, visits to $s_{\frac{N}{2}}$ are finally suppressed and the agent converges to the locally optimal policy in $s_1$. See Appendix for the comparison of visit counts for other sets of $N$ and for NoisyNet.

\begin{figure}[h]
\centering
\subfigure[Variational DQN]{\includegraphics[width=.43\linewidth]{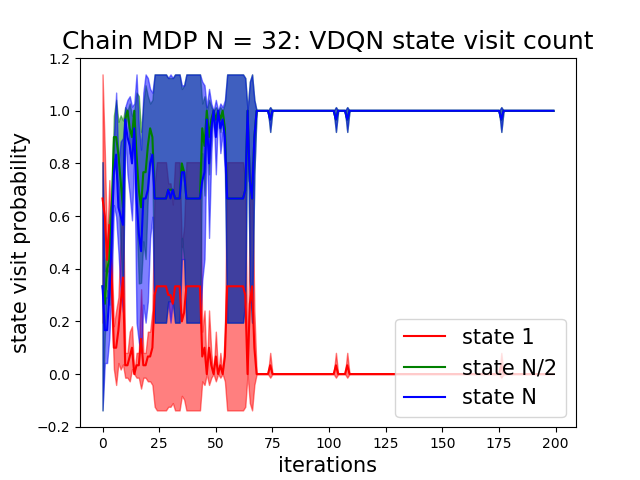}}
\subfigure[DQN]{\includegraphics[width=.43\linewidth]{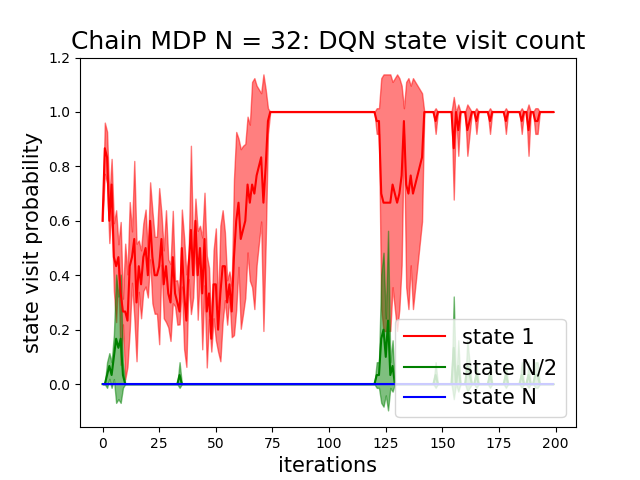}}
\caption{Variational DQP vs. DQN in Chain MDP $N=32$: state visit counts. Count $c_n = 1$ for state $1\leq n\leq N$ if state $s_n$ is ever visited in one episode. Running averages over multiple consecutive episodes of $c_n$ produces $p_n$, which is an approximate state visit probability under current policy. Each probability curve $p_n$ is average over multiple initializations. Each iteration is $10$ episodes.}
\end{figure}

\newpage
\section{Conclusion}
We have proposed a framework to directly tackle the distribution of the value function parameters. Assigning systematic randomness to value function parameters entails efficient randomization in policy space and allow the agent to do efficient exploration. In addition, encouraging high entropy over parameter distribution prevents premature convergence. We have also established an equivalence between the proposed surrogate objective and variational inference loss, which allows us to leverage black box variational inference machinery to update value function parameters.

Potential extension of our current work will be to apply similar ideas to Q-learning in continuous control tasks and policy gradient methods. We leave this as future work.

\small

\bibliographystyle{apa}
\bibliography{your_bib_file.bib}

\newpage
\section{Appendix}
\subsection{Derivation of Variational Inference Interpretation}
Consider a bayesian network $Q_\theta(s,a)$ with input $(s,a)$ and parameter $\theta$. The parameter $\theta$ has a prior $p(\theta)$. This bayesian network produces mean for a Gaussian distribution with variance $\sigma^2$ i.e. let $d_j$ be a sample
$$d_j|\theta \sim N(Q_\theta(s,a),\sigma^2)$$
Given $N$ samples $D = \{d_j\}_{j=1}^N$, the posterior $p(\theta|D)$ is in general not possible to evaluate.
Hence we propose a variational family distribution $q_\phi(\theta)$ with parameter $\phi$ to approximate the posterior. Variational inference literature \cite{blei2015,blei2017} has provided numerous techniques to compute $\phi$, yet for a flexible model $Q_\theta(s,a)$ black box variational inference is most scalable. We consider minimizing KL divergence between $q_\phi(\theta)$ and $p(\theta|D)$ 
\begin{align}
\mathbb{KL}[q_\phi(\theta) || p(\theta|D)] = \mathbb{E}_{\theta\sim q_\phi(\theta)}\big[\log \frac{q_\phi(\theta)}{p(\theta)p(D|\theta)}\big] + \log p(D)
\label{eq:kldecompose}
\end{align}
Let $p(\theta)$ be improper uniform prior. Also recall that $$\log p(D|\theta) = \sum_{j=1}^N p(X|\theta) = -\sum_{j=1}^N \frac{(Q_\theta(s_j,a_j) - d_j)^2}{2\sigma^2} + N \log (\frac{1}{\sqrt{2\pi}}\frac{1}{\sigma})$$ 

Decompose the objective in (\ref{eq:kldecompose}) and omit constants, we get
\begin{equation}
\mathbb{E}_{\theta\sim q_\phi(\theta)}\big[\log q_\phi(\theta)\big] + \mathbb{E}_{\theta\sim q_\phi(\theta)} \big[\sum_{j=1}^N \frac{(Q_\theta(s_j,a_j) - d_j)^2}{2\sigma^2}\big]
\end{equation}

We then identify the first term as $-\mathbb{H}(q_\phi(\theta))$ and the second as expected Bellman error.

\subsection{Variational Inference as approximate minimization of Bellman error}
Assume the Bayesian network $Q_\theta(s,a)$ produces Gaussian sample $d_j |\theta \sim N(Q_\theta(s,a),\sigma^2)$. Given samples $D = \{d_j\}_{j=1}^N$, as $N \rightarrow \infty$
$$p(\theta|D) \approx \arg\min_{q(\theta)}\mathbb{E}_{\theta \sim q(\theta)} \big[\frac{1}{N}\sum_{j=1}^N (Q_\theta(s_j,a_j) - d_j)^2\big]$$
as the information in prior $p(\theta)$ is overwhelmed by data. In fact, $p(\theta|D)$ itself will also converge to MAPs. Using variational family distribution $q_\phi(\theta)$ to approximate the posterior we expect
$$q_\phi(\theta) \approx \arg\min_{q(\theta)} E_{\theta\sim q(\theta)} \big[\frac{1}{N}\sum_{j=1}^N (Q_\theta(s_j,a_j) - d_j)^2\big]$$
therefore any variational inference procedure that updates $q_\phi(\theta)$ to approximate the posterior $p(\theta|D)$ will converge to an approximate minimizer of the Bellman error. In particular, variational inference using $\mathbb{KL}[q_\phi(\theta)||p(\theta|D)]$ will result in an additional entropy bonus term in the surrogate objective, which in effect encourages dispersed policy distribution and efficient exploration.

\subsection{Implementation Details}
Since we have established an equivalence between the surrogate objective (Bellman error $+$ entropy bonus) and variational inference loss, we could leverage highly optimized implementation of probabilistic programming packages to perform parameter update. In our experiment, we used Edward \cite{tran2017} to minimize the KL divergence between variational distribution $q_\phi(\theta)$ and the posterior $p(\theta|D)$.

In classic control tasks, we train Variational DQN, DQN and NoisyNet agents for about $800\sim 1000$ episodes on each task. The learning rate is $\alpha = 10^{-3}$ or $\alpha = 10^{-2}$. The discount factor is $\gamma = 0.99$. 

In Chain MDPs, we train all  agents for about $2000$ episodes for fixed $N$. The learning rate is $\alpha = 10^{-3}$ or $\alpha = 10^{-2}$. The discount factor is $\gamma = 1.0$ (if discount factor $\gamma < 1$ is used, when $N$ is large, going to state $s_1$ will be optimal). 

For all experiments, the batch size of each mini-batch sampled from the replay buffer is $64$. The target network is updated every $100$ time steps.  Each experiment is replicated multiple times using different random seeds to start the entire training pipeline. DQN uses an exploration constant of $\epsilon = 0.1$ throughout the training. Variational DQN and NoisyNet both use component-wise gaussian distribution to parameterize distribution over value function parameters, and they are all initialized according to recipes in \cite{fortunato2017}. Variational DQN is updated using KLqp inference algorithm \cite{blei2017} with regularization constant $\lambda = 2 \cdot 10^{-2}$. 

\subsection{Recover DQN}
We set the variational family distribution to be point distribution $\theta = \delta(\phi)$. Hence $\phi$ has the same dimension as $\theta$ and is in effect $\theta$ itself. Then we apply Maximum a Posterior (MAP) inference to update $\phi$. Under Gaussian generative model of $Q_\theta(s,a)$ and improper uniform prior, this is equivalent to minimizing Bellman error only.

\subsection{Chain MDP: Visit Count for $N=8,32$ and $N=128$}

Below we present the state visit counts for $N = 8,32$ and $128$ for all three algorithms (Variational DQN, DQN and NoisyNet). For small $N$ ($N = 8$), Variational DQN and NoisyNet both identify the optimal path faster yet display larger variance in performance, while DQN makes progress in a more steady manner. 

For medium sized $N$ ($N = 32$), Variational DQN still manages to converge to the optimal policy though the initial exploration stage exhibits larger variance. DQN occasionally pass the middle point $s_{\frac{N}{2}}$ (observed from the green spike) but cannot reach $s_N$. NoisyNet explores more efficiently than DQN since it sometimes converge to optimal policy but is less stable than Variational DQN.

For large $N$ ($N=128$), Variational DQN takes more time to find the optimal policy but still converges within small number of iterations. On the other hand, both DQN and NoisyNet get stuck.

\begin{figure}[h]
\centering
\subfigure[Variational DQN $N=8$]{\includegraphics[width=.3\linewidth]{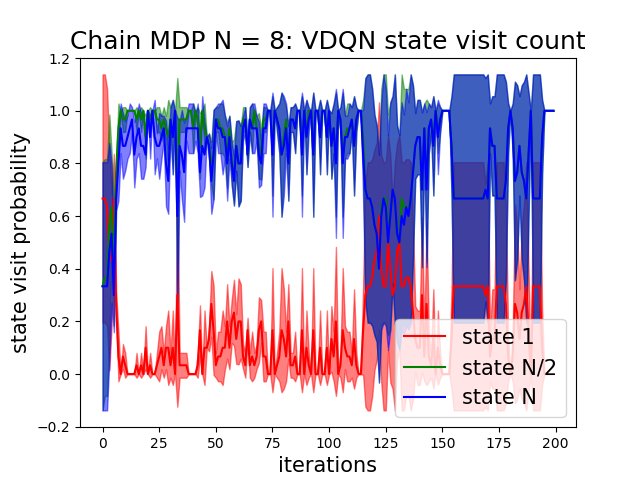}}
\subfigure[Variational DQN $N=32$]{\includegraphics[width=.3\linewidth]{VIDQN_MDPN32visit}}
\subfigure[Variational DQN $N=128$]{\includegraphics[width=.3\linewidth]{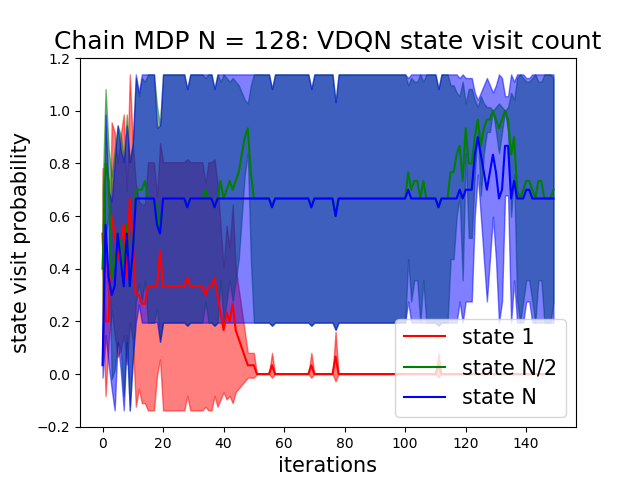}}
\subfigure[DQN $N=8$]{\includegraphics[width=.3\linewidth]{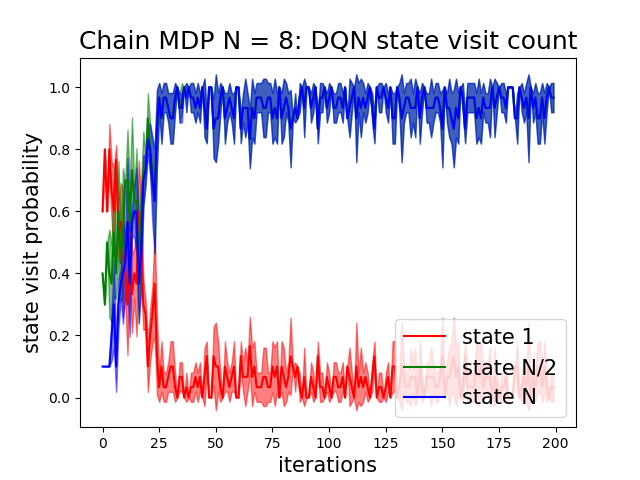}}
\subfigure[DQN $N=32$]{\includegraphics[width=.3\linewidth]{DQN_MDPN32visit}}
\subfigure[DQN $N=128$]{\includegraphics[width=.3\linewidth]{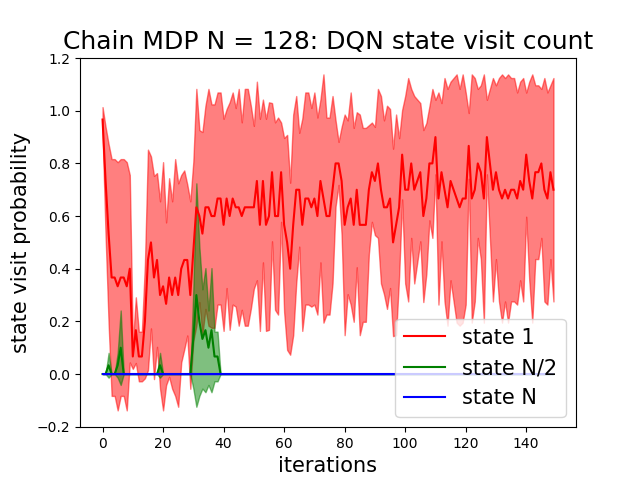}}
\subfigure[NoisyNet $N=8$]{\includegraphics[width=.3\linewidth]{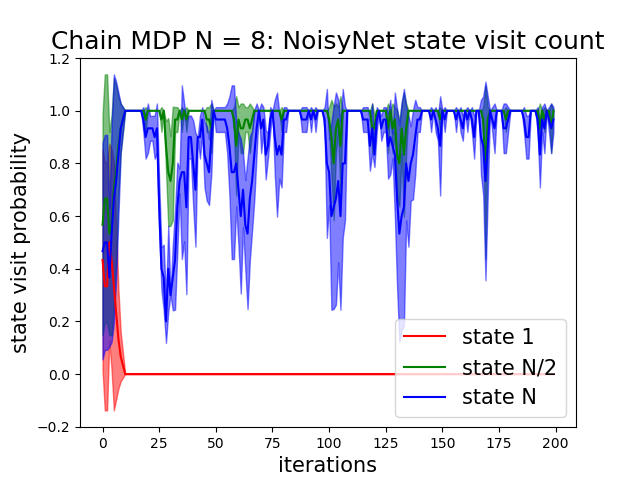}}
\subfigure[NoisyNet $N=32$]{\includegraphics[width=.3\linewidth]{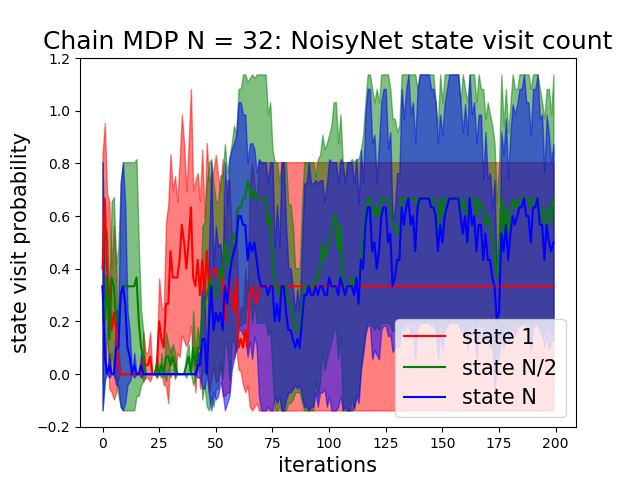}}
\subfigure[NoisyNet $N=128$]{\includegraphics[width=.3\linewidth]{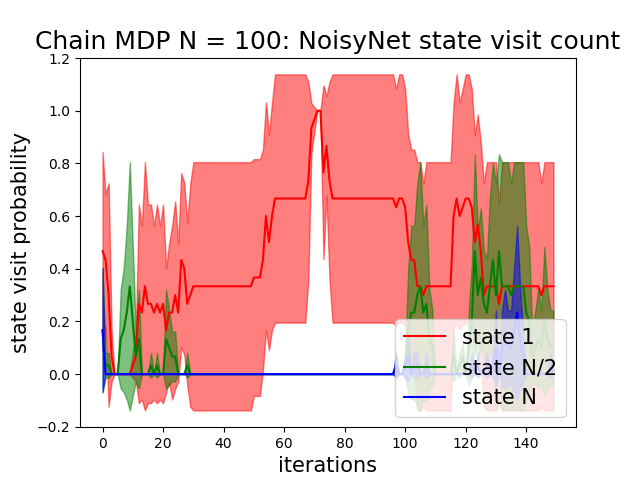}}
\caption{Variational DQP vs. DQN vs. NoisyNet in Chain MDP $N=8,128$: state visit counts. Count $c_n = 1$ for state $1\leq n\leq N$ if state $s_n$ is ever visited in one episode. Running averages over multiple consecutive episodes of $c_n$ produces $p_n$, which is an approximate state visit probability under current policy. Each probability curve $p_n$ is average over multiple initializations. Each iteration is $10$ episodes.}
\end{figure}

\subsection*{Interpretation as Approximate Thompson Sampling}
Thompson sampling \cite{william1933} is an efficient exploration scheme in multi-arm bandits and MDPs. At each  step, Variational DQN maintains a distribution over action-value function $Q_\theta(s,a)$. The algorithm proceeds by sampling a parameter $\theta\sim q_\phi$ and then select action $a_t = \arg\max_a Q_\theta(s_t,a)$. This sampling differs from exact Thompson sampling in two aspects.
\begin{itemize}
\item The sampling distribution $q_\phi$ is not the posterior distribution of $\theta$ given data but only a variational approximation.
\item During mini-batch training, the data $D$ used to update the posterior $p(\theta|D)$ is not generated by exact action-value function $Q^\ast(s,a)$ but the target network $Q_{\theta^{-}}(s,a)$.
\end{itemize}
Hence the quality of this exact Thompson sampling depends on the quality of both approximation $q_\phi(\theta)\approx p(\theta|D)$ and $Q_{\theta^{-}}(s,a) \approx Q^\ast(s,a)$, the second of which is in fact our very goal. When the approximation is not perfect, this approximate sampling scheme can still be beneficial to exploration.

\end{document}